\documentclass[journal]{vgtc}                

\ifpdf
  \pdfoutput=1\relax                   
  \usepackage{graphicx}                
  \DeclareGraphicsExtensions{.pdf,.png,.jpg,.jpeg} 
\else
  \usepackage{graphicx}                
  \DeclareGraphicsExtensions{.eps}     
\fi%

\graphicspath{{figures/}{pictures/}{images/}{./}} 

\usepackage{microtype}                 
\PassOptionsToPackage{warn}{textcomp}  
\usepackage{textcomp}                  
\usepackage{mathptmx}                  
\usepackage{times}                     
\usepackage{cite}                      
\usepackage{tabu}                      
\usepackage{booktabs}                  

\usepackage{graphicx}
\usepackage{subfigure}
\usepackage{float}
\usepackage{capt-of}
\usepackage{amssymb}
\usepackage{rotating}
\usepackage{multirow}
\usepackage{bm}
\usepackage{hhline}
\usepackage{color,soul}

\usepackage{graphicx}
\usepackage{setspace}
\usepackage{amsmath}
\usepackage{bm}
\usepackage{subfigure}
\usepackage{tikz}
\usepackage{amssymb}
\usepackage{pifont}
\usepackage{hyperref}
\usepackage{url}

\DeclareMathOperator*{\argmin}{arg\,min}

\onlineid{1072}

\vgtccategory{Research}
\vgtcpapertype{algorithm/technique}

\title{
Instant Visual Odometry Initialization for Mobile AR
}

\author{Alejo Concha, Michael Burri, Jes\'us Briales, Christian Forster, and Luc Oth}

\authorfooter{

All authors are with Facebook Zurich, Switzerland. E-mail:  {\tt\small aconchabelenguer@gmail.com  alejocb@fb.com}

}

\shortauthortitle{Biv \MakeLowercase{\textit{et al.}}: Instant Visual Odometry Initialization for Mobile AR}

\abstract{Mobile AR applications benefit from fast initialization to display world-locked effects instantly. However, standard visual odometry or SLAM algorithms  require motion parallax to initialize (see Figure \ref{fig:teaser}) and, therefore, suffer from delayed initialization. In this paper, we present a 6-DoF monocular visual odometry that initializes instantly and without motion parallax. Our main contribution is a pose estimator that decouples estimating the 5-DoF relative rotation and translation direction from the 1-DoF translation magnitude. While scale is not observable in a monocular vision-only setting, it is still paramount to estimate a \emph{consistent} scale over the whole trajectory (even if not physically accurate) to avoid AR effects moving erroneously along depth. In our approach, we leverage the fact that depth errors are not perceivable to the user during rotation-only motion. However, as the user starts translating the device, depth becomes perceivable and so does the capability to estimate consistent scale. Our proposed algorithm naturally transitions between these two modes. 
Our second contribution is a novel residual in the relative pose problem to further improve the results. The residual combines the Jacobians of the functional and the functional itself and is minimized using a Levenberg–Marquardt optimizer on the 5-DoF manifold. We perform extensive validations of our contributions with both a publicly available dataset and synthetic data. We show that the proposed pose estimator outperforms the classical approaches for 6-DoF pose estimation used in the literature in low-parallax configurations. Likewise, we show our relative pose estimator outperforms  state-of-the-art approaches in an odometry pipeline configuration where we can leverage initial guesses. We release a dataset for the relative pose problem using real data to facilitate the comparison with future solutions for the relative pose problem. Our solution is either used as a full odometry or as a pre-SLAM component of any supported SLAM system (ARKit, ARCore) in world-locked AR effects on platforms such as Instagram and Facebook.
} 


\CCScatlist{ 
 \CCScat{K.6.1}{Management of Computing and Information Systems}%
{Project and People Management}{Life Cycle};
 \CCScat{K.7.m}{The Computing Profession}{Miscellaneous}{Ethics}
}

\vgtcinsertpkg

\begin{document}


 
\firstsection{Introduction}

\maketitle

Knowing the precise 6-DoF motion of a mobile phone allows us to augment the real-world with virtual effects. A popular use-case of this capability is the placement of virtual furniture to make purchasing decisions. The pose of the mobile phone can be estimated with camera-based Simultaneous Localization and Mapping (SLAM) or Visual Odometry (VO) -- Cadena et al. \cite{cadena2016past}. These methods detect and track salient features in the environment and thereby localize the camera. While SLAM builds and optimizes a map of the observed scene, a VO system just maintains a sliding-window of the most recent estimated camera poses and landmarks at the cost of lower global accuracy but higher compute efficiency. Since many world-locked AR experiences are very short in nature, and compute usage is of high importance, a VO system suffices for most applications. 

\begin{figure}[ht]
  \centering
  \includegraphics[width=0.99\linewidth]{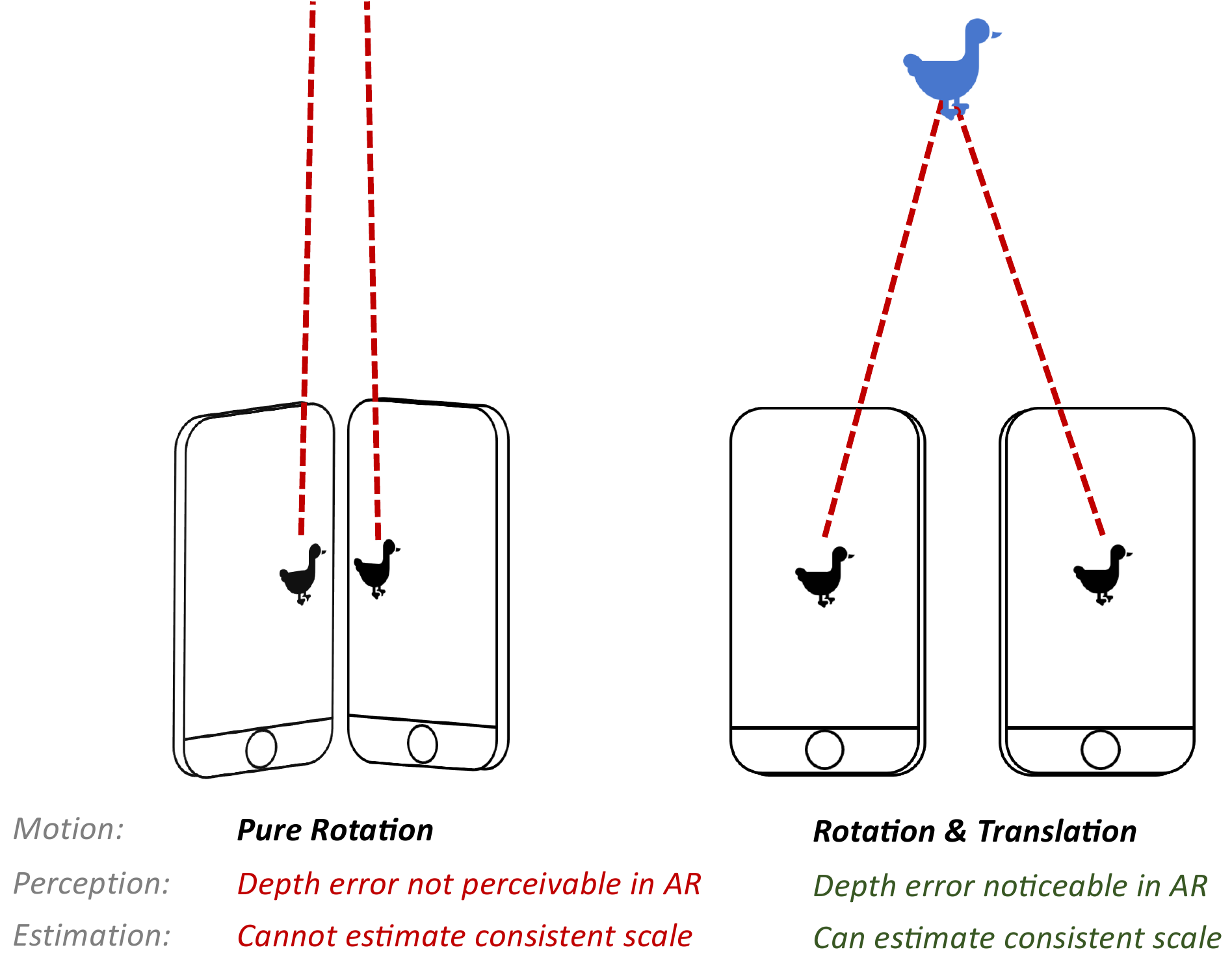}
  \caption{The left part of the Figure shows that depth is neither perceivable nor recoverable during pure rotational motion. On the other hand, if the device translates as shown on the right, it is possible to estimate consistent depth in a monocular setting up to a constant scale factor. However, an in-consistent scale over multiple frames would be user noticeable. Our initialization method naturally moves from the left to the right by starting to estimate translation magnitude corresponding to a consistent scale as depth errors become perceivable.}
  \label{fig:teaser}
\end{figure}

Widely used mobile SDKs such as ARKit or ARCore leverage the inertial measurement unit (IMU) in addition to the camera sensors. The IMU is an ideal complementary sensor to the camera as it (1) renders gravity and scale observable and (2) provides reliable angular velocity and linear acceleration even when the cameras are occluded or the images suffer motion-blur or low-contrast. However, there exists a large body of devices without accurate and stable spatial-temporal camera-IMU calibration. Additionally, there exist many low-end smartphones with accelerometer but no gyroscope or phones with faulty sensor data due to manufacturing imprecisions. Ubiquitous mobile AR needs to support low-end devices, which motivated the development of our vision-only VO system. Due to the major advantage of leveraging the IMU, the proposed system is able to optionally include and process inertial measurements in case they are available.

Many world-locked mobile AR experiences are very short in nature (a few seconds) as users rapidly browse through different AR effects that rely on different underlying capabilities such as world-locked, face-locked, and object-locked. One way to increase user satisfaction is to initialize the tracker instantaneously to minimize time-to-fun. However, standard VO and visual-inertial odometry initialization techniques require translational motion to initialize (approx. 10 cm, depending on the scene depth). This is because many closed-form solutions that are used to jointly initialize the initial camera poses and 3D landmarks require translation to triangulate the 3D points or may even be degenerate in the rotation-only case. This motivated our work on a VO system that initializes instantaneously and independently of the user motion. Our solution is either used as a full odometry or as a pre-SLAM component of any supported SLAM system (ARKit, ARCore) in world-locked AR effects on platforms such as Instagram and Facebook.

While our contribution has been motivated by the mobile AR use-case, there are several other applications in the robotics community that rely on instant initialization of VO or SLAM. An example is the rapid deployment of vision-based micro aerial vehicles as shown in Faessler et al. \cite{faessler2015automatic}.


\section{Related work}
 
Initializing stereo (Mur-Artal et al. \cite{mur2017orb}) or RGB-D systems (Newcombe et al. \cite{newcombe2011kinectfusion}, Concha and Civera \cite{concha2017rgbdtam}) is straightforward in the vast majority of cases. These systems can estimate the depth of features without moving the device. Other works rely on IMU sensors for fast initialization, like Bloesch et al. \cite{bloesch2015robust} or Li et al. \cite{li2019rapid}. Structural cues might also be used as in Huang et al. \cite{huang2017fast}, which propose a method for initializing from first frame using vanishing points and some indoor cues. Moreover, deep-learning-based depth estimation has the potential to initialize a map from first frame (F\'acil et al. \cite{facil2019cam}).

In this paper, and contrary to some of the above-mentioned works, we do not make any assumptions regarding the structure of the environment and we do not use additional sensors to solve this problem either. In this context, VO initialization can be divided in two main research lines: feature-based (indirect) initialization methods and direct initialization methods. In the next section, we discuss different initialization techniques that have been explored in both research lines:

\subsection{Feature-based (indirect) initialization methods}
Feature-based initialization approaches extract features in an image that are matched in another image of the same scene with a different viewpoint. Feature matches are used to estimate both an initial relative pose between the views and a map of 3D features using either closed-form solvers or optimization-based techniques:

\subsubsection{Minimal solvers based on Essential and Homography matrices}
\label{essential}
The relative pose can be recovered by estimating an Essential matrix (eight-point algorithm, Zisserman and Hartley \cite{andrew2001multiple}) or a Homography matrix (Faugeras and Lustman \cite{faugeras1988motion}, Longuet  \cite{longuet1986reconstruction}) between the cameras, depending on the camera motion and scene structure. Camera poses are then used to triangulate the feature correspondences and build a map that is used for tracking in the subsequent frames. Even though these solvers have been used successfully for decades, they still present some issues that motivated this paper:

\begin{itemize}
\item Non-instantaneous initialization. The Essential matrix might not be correctly estimated in the case of zero translation. A zero matrix becomes a valid solution in such situations and therefore its constraints deteriorate.
\item Non-minimal parametrization: The parameters from the Essential and the Homography matrix are not directly related to the actual motion-related parameters and the parametrization is not minimal, which makes it harder to withdraw conclusions from the actual estimated values during the optimization.
\item Structure-dependency. Essential matrix estimation is an ill-defined problem with planar scenes. A Homography model (Faugeras et al. \cite{faugeras1988motion}) is proposed in the literature for these situations as backup plan (Mur-Artal et al. \cite{mur2015orb}). A Homography model cannot be applied to non-planar scenes, unless points are far away where translation cannot be recovered. Mur-Artal et al. \cite{mur2015orb} deal with this problem by using a model selection strategy: An homography and an Essential matrix are estimated in parallel and the one that produces the lowest re-projection error is taken as a candidate for initialization. 
\item Solution multiplicity. Homography solvers return multiple solutions (Faugeras et al. \cite{faugeras1988motion}) that need to be disambiguated. Some of the solutions can be trivially rejected but for others there is not a better option than triangulate and track feature correspondences for a few frames with a SLAM system to verify their re-projection error in subsequent frames Mur-Artal et al. \cite{mur2015orb}.
\end{itemize}

\subsubsection{Optimization-based initialization methods}
Several optimization-based approaches have been explored that address some of the limitations of minimal solvers that estimate an Essential or Homography matrix. Kneip and Lynen \cite{kneip2013direct} have addressed this problem proposing a direct optimization of frame to frame rotation. Lee and Civera \cite{lee2021rotation} have recently extended this work to multiple views, proposing a rotation-only bundle adjustment optimizer.
%
On the downside, these approaches need to cope with non-trivial and non-convex optimization problems for which finding the right optimal solution vs other minima can turn challenging.
Briales et al. \cite{briales2018certifiably} proposed for the first time a solver for (an equivalent form of) the problem formulation by Kneip and Lynen \cite{kneip2013direct} which comes with global optimality guarantees. The approach however relies on solving a convex SDP relaxation of an equivalent Quadratically Constrained Quadratic Program (QCQP) formulation of the problem (Park and Boyd \cite{park2017general}), which is computationally intensive (with respect to real-time standards) and makes it non-straightforward to leverage initial guesses for the relative pose. Garc\'ia-Salguero et al. \cite{garcia2021certifiable} extends this work and proposes an optimality certifier for the relative pose problem.

The main advantage of relative pose estimators is their ability to accurately estimate rotations independently of the scene structure and motion of the camera. Translations are also independent of the scene structure and are accurately estimated if there is enough baseline between the cameras. For these reasons, we have used a relative pose estimator in our formulation, see Section \ref{contributions} for more details.
 
\subsection{Direct initialization methods}
\label{direct}
Direct approaches directly use pixel intensity for tracking and mapping. For initialization, direct approaches either use the above-mentioned solvers from indirect approaches to initialize their systems (Engel et al. \cite{engel2014lsd}) or create an inaccurate first map. For example, in Engel et al. \cite{engel2017direct}, a random map is initialized and in Concha and Civera \cite{concha2015dpptam} the features are placed at a constant distance from the first camera. In these approaches, the initial inaccurate map is continuously refined as the camera moves hoping for an eventual convergence of the map parameters to their true values. Generating an initial inaccurate map can work in some situations but can fail catastrophically if the initial map is very different from the actual one. This is because the map is directly used to estimate all 6-DoF parameters, meaning errors in the map are directly propagated to all estimation parameters and the estimation can eventually diverge. 

\section{Contributions and Outline}
\label{contributions}
Optimization-based relative pose estimators solve most of the limitations (section \ref{essential}) that are present in classical minimal solvers that estimate the Essential matrix between two views. Even though relative pose solvers are good at estimating the two-view problem, they fail to estimate a \emph{consistent} scale in a VO setting as they return a unit-norm translation. While the \emph{true} scale is not observable in a monocular vision-only setting, it is still paramount to maintain a consistent scale by estimating the translation magnitude in the two-view problem. This is required to recover a camera trajectory that is accurate up to a single scale factor that remains constant throughout the session. To solve this issue:

\begin{itemize}
\item  We propose to combine a relative pose estimator with a translation-magnitude-estimator. The translation-magnitude-estimator minimizes keyframe to frame re-projection errors using depth-estimates of past feature observations as input. With this formulation, we use all feature correspondences to estimate the relative pose while only using the correspondences with estimated depth to estimate the magnitude of the translation. In traditional SLAM systems, only those features with an accurate estimated depth can be used to estimate the full 6-DoF pose of the camera. This is why direct approaches that initialize an inaccurate map can fail during initialization when the map is inaccurate (see Section \ref{direct}), the error in the map is directly propagated to all motion parameters which can eventually diverge. We also initialize an initial inaccurate map, however, in our case the errors in the map do not affect the estimated relative pose but only the estimated translation magnitude. We exploit the fact that while translation is not observable, the accuracy of the map does not matter. The reverse also holds: at the point where translation becomes significant, feature depth becomes observable.  We show in the experimental Section \ref{experiments} that our 5-DoF + 1-DoF optimizer outperforms the classical 6-DoF pose estimator in low-parallax motions.
\item One challenge with solving the optimization problem underlying relative pose estimators is that the underlying formulation is not a sum of squares, but rather a sum of algebraic errors. As a result we cannot directly apply classical simple and lightweight Gauss-Newton-like algorithms, but need to resort to more generic Newton-like methods like Riemannian Trust Regions -- Absil et al. \cite{absil2009optimization}. These more generic solvers are not as generally available as classical Least Squares solvers like Gauss-Newton or Levenberg–Marquardt (More \cite{more1978levenberg}), and often lack the maturity of the latter (specially when it comes to availability as efficient C++ implementations within common libraries).
Besides, in resource-constrained platforms like mobile devices, where each new library dependency matters to minimize binary size, it is desirable that we rely on a classical Least Squares solver.

As an alternative to the above-mentioned limitations, Kneip and Lynen \cite{kneip2013direct} propose to minimize a different functional consisting of the sum of squared Jacobians of the original functional and therefore a classical Gauss-Newton or Levenverg-Marquardt algorithm can be applied. 
This approach, by definition, is equivalent to finding a local minimum for the original functional (where Jacobians become null). The surrogate residual solved in Kneip and Lynen \cite{kneip2013direct} has multiple equally valid global minimum (as many as local minimum exist in the original functional), making it much harder to converge to the globally optimal solution.

As a second contribution, we extend the Jacobian minimization trick of Kneip and Lynen \cite{kneip2013direct} and include the objective function itself in the residual. This basically makes the surrogate residual well-defined as there will be a single global optimum, making convergence to the right solution much easier. Another characteristic of this approach is that it allows the usage of initial guesses for the parameters to estimate, which is a big advantage for visual odometries.
\end{itemize}
 
The rest of the paper is structured as follows: the problem is defined in Section \ref{definition}. Sections \ref{solution} and \ref{details} explain the proposed approach in detail. We validate our contributions in Section \ref{experiments}. Finally, conclusions and future lines of work are given in Section \ref{conclusion}. Additionally, we include an Appendix in Section \ref{appendix} for the derivations omitted from Section \ref{solution}.

\section{Problem definition}
\label{definition}

Our goal is to estimate in real-time and without initialization a consistent 6-DoF trajectory of the motion of a monocular device with calibrated camera parameters. We can solve this by estimating  $\bm{T}\in \tt{SE}(3)$, the pose transform between a previous frame (keyframe) and the current frame given a set of feature correspondences between them ($\bm{f_i}$ in the past keyframe and $\bm{f_i}^\prime$ in the current frame) and -- optionally -- the depth of the features in the source view ($d_i$). Where $i$ denotes the index of the feature.  Note if $\bm{p_i}$ is a 3D point in the reference frame of the first view and $\bm{p_i}^\prime$ is the same point in the reference frame of the second view, they are related through $\bm{T}$ as $\bm{p_i}^\prime = \bm{T}\bm{p_i}$.
\newline
Where $\bm{T}$ is defined as follows:

\begin{equation} 
\label{parametrization}
\centering
\bm{T} = \begin{bmatrix} \bm{R} & \bm{u} s \\ {0}_{1\times 3} & 1\end{bmatrix}, \text{s.t. } \bm{R} \in \text{SO(3)}, \bm{u} \in S^2, s \in \mathbb{R}
\end{equation}
\normalsize

Where $\text{SO(3)}$ is the rotation group (Stuelpnagel \cite{stuelpnagel1964parametrization}) and $S^n$ represents points in the n-dimensional unit sphere. $\bm{u}$ is the translation direction, a 3D unit-norm vector.

Through the paper, we refer to the magnitude of the translation as $s \in \mathbb{R}$. Whenever we refer to 5-DoF in this paper we mean the unscaled relative pose between two views ($\bm{R},\bm{u}$), while if we refer to 6-DoF we mean we additionally estimate the magnitude of the translation to produce a full 6-DoF pose ($\bm{R}, \bm{u}, s$) $\rightarrow$ ($\bm{R}, \bm{t}$)  with consistent scale along the trajectory (but \emph{not} metric scale, as this is a monocular odometry).
 

\section{6-DOF Pose estimation}
\label{solution}

\begin{figure*}[ht!]
\centering
     \includegraphics[width=1.0\textwidth]{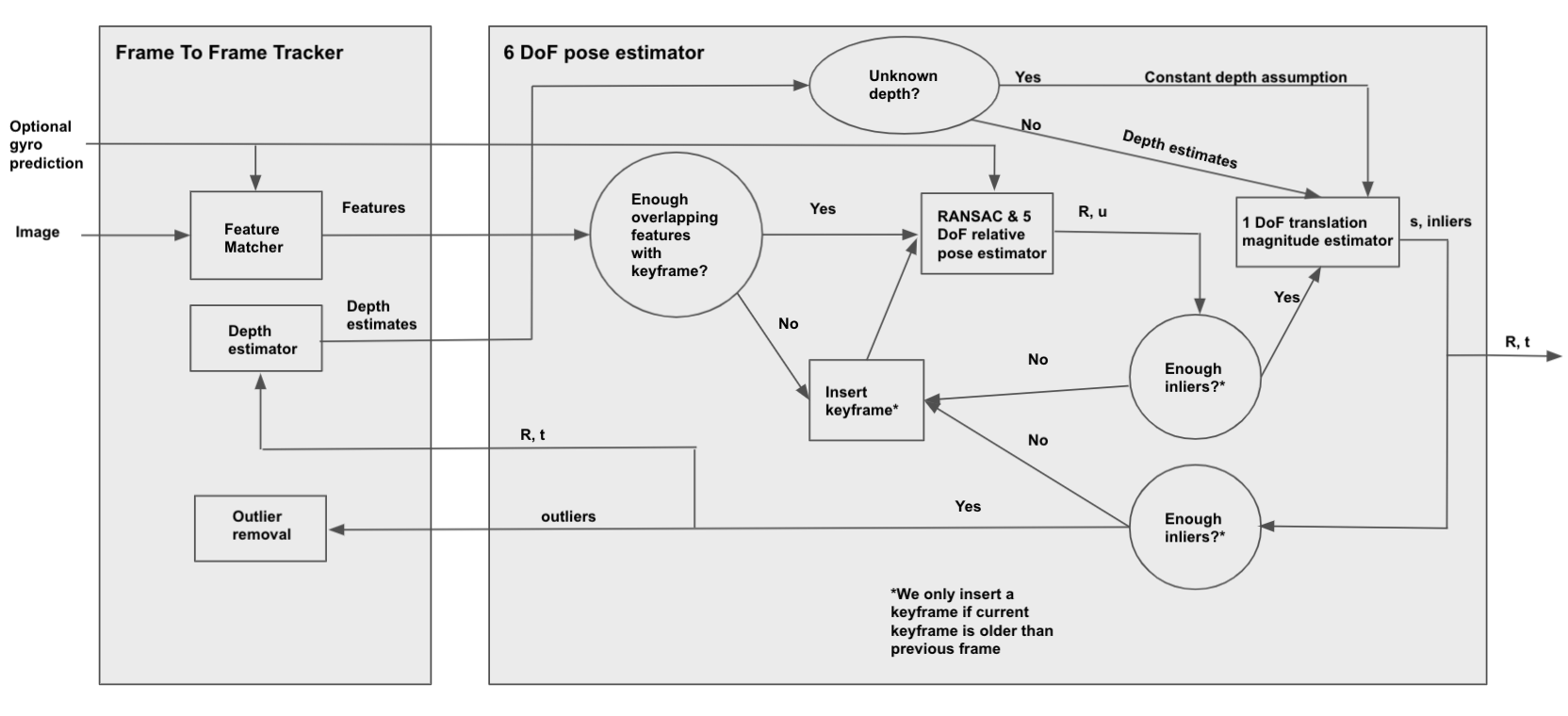}
      \caption{Pipeline of our approach. Images (and optionally IMU gyro predictions) feed our visual odometry. The frame-to-frame tracker matches features and produces feature correspondences (bearing vectors) that are consumed by the 6-DoF pose estimator. The pose estimator contains a relative pose estimator (5-DoF) that takes the bearing correspondences and estimates a 5-DoF pose. The 5-DoF pose and the estimated depth of the features (depth is estimated by the tracker) are used to estimate the magnitude of the translation, completing the final 6-DoF pose. If depth is not available (during the first frames), we use the constant depth assumption. A new keyframe is inserted in the previous frame if there are not enough overlapping features between current keyframe and last frame or if any of the estimators fail. If current keyframe is already the previous frame, we do not insert a new keyframe. Outliers from pose estimators are sent to the tracker for removal.}
       \label{fig:pipeline}
\end{figure*}

Our pipeline is composed of two main components, a frame to frame tracker and a pose estimator (see Figure \ref{fig:pipeline}). In this section, we explain the proposed 6-DoF estimator, which is the main contribution of this paper. We will explain the implementation details of the frame-to-frame tracker in the next section for completeness.

Following the 6-DoF parametrization in eq. \ref{parametrization}, we can split the 6-DoF estimation problem into (1) the estimation of the relative pose ($\bm{R} \in \text{SO(3)}$ with unit-norm translation $\bm{u} \in S^2$) and (2) the estimation of the magnitude of the translation ($s$). The relative pose can be estimated using all feature correspondences between two views using a relative pose estimator (see Section \ref{relative} for details). Once the relative pose is estimated, we estimate the magnitude of the translation. To estimate the magnitude of the translation, we also need as input the depth of the \emph{optionally} triangulated features, which might have been estimated using past 6-DoF poses via triangulation (see Section \ref{scale} for details). 
   
\subsection{5-DoF Relative Pose Estimator}
\label{relative}

The relative pose problem is the problem of determining the relative rotation and direction of the translation between two frames using 2D-2D correspondences.

\begin{figure}[bt]
\centering
\includegraphics[width=0.460\textwidth]{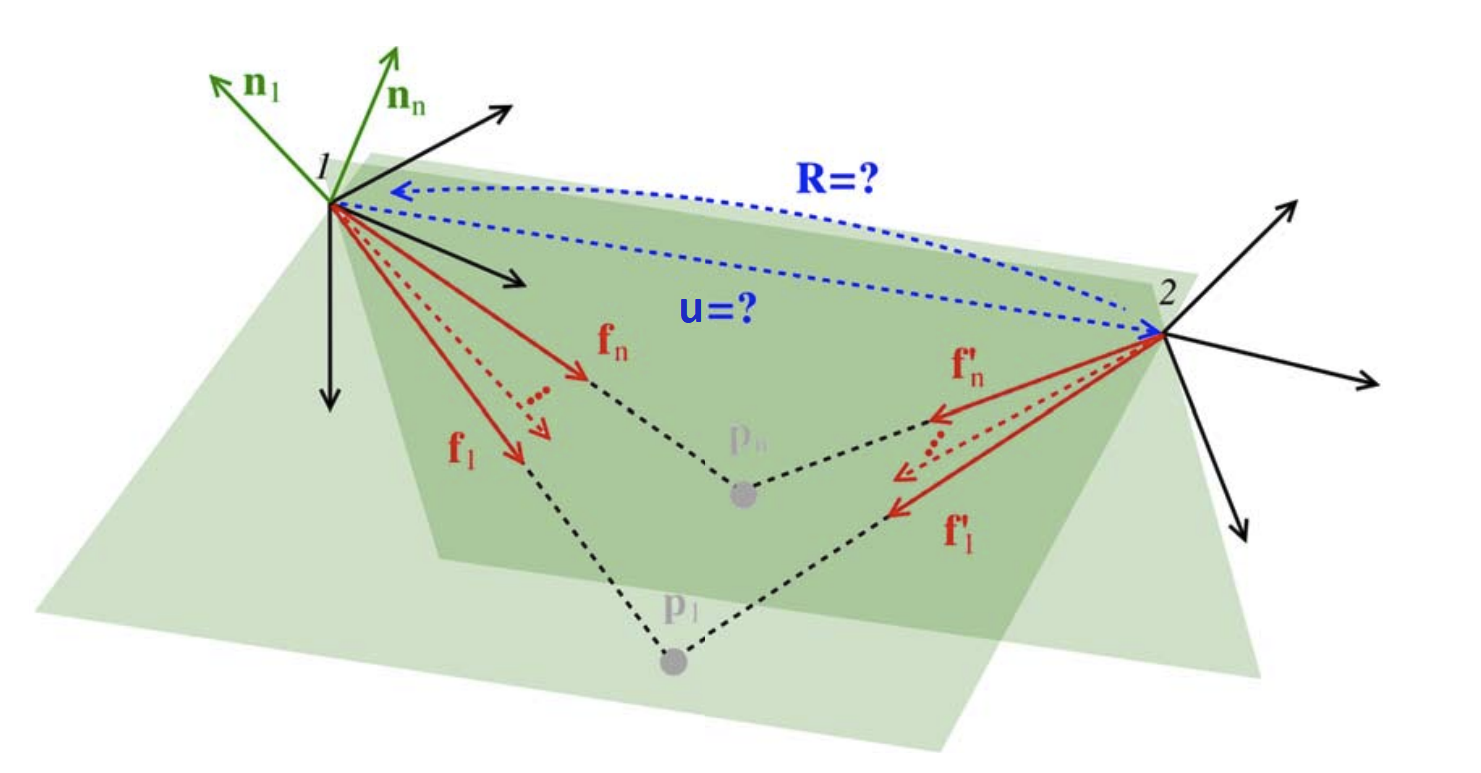}
\caption{Graphical definition of the relative pose problem. Bearing vectors (in red) from two different views are the known variables. The rotation $\bm{R}$ and translation direction  $\bm{u}$ between the views are estimated by minimizing the equation \ref{eq:final3}. Figure borrowed from Kneip and Lynen \cite{kneip2013direct}.}.
\label{fig:kneip_fig}
\end{figure}
  
Given the bearing vectors ($\bm{f_i}$ and $\bm{f_i}^\prime$) from a set of correspondences, Kneip and Lynen \cite{kneip2013direct} proposed to solve the relative pose problem by enforcing the co-planarity of the epipolar plane normals $\bm{m_i} =  \bm{R} \bm{f_i} \times \bm{f_i}^\prime$ for all epipolar planes (see Figure \ref{fig:kneip_fig} for a graphical explanation). This is achieved by minimizing the minimum eigenvalue of the covariance matrix of the epipolar plane normals $\bm{M}(\bm{R}) = \sum_{i=1}^{n}  \bm{m_i} \bm{m_i}^\intercal $, where the rotation matrix $\bm{R}$ is the variable to optimize:

\begin{equation} 
\hat{\bm{R}} = \argmin_{\bm{R}} \lambda_{min}\left( 
\bm{M\left(\bm{R}\right)} \right)
\end{equation}

We refer the reader to the original work Kneip and Lynen \cite{kneip2013direct} for more details on the derivation of the functional. Once $\bm{R}$ is estimated, $\bm{u}$ can be retrieved as the eigenvector corresponding to the minimum eigenvalue of $\bm{M\left(\hat{\bm{R}}\right)}$:

\begin{equation} 
\hat{\bm{u}} = \argmin_{\bm{u}} \bm{u} \bm{M\left(\hat{\bm{R}}\right)} \bm{u}^\intercal
\label{eq:t_ab}
\end{equation}

Instead, we follow a similar approach proposed by Briales et al. \cite{briales2018certifiably}, solving translation and rotation simultaneously. The equation \ref{eq:t_ab} is minimized directly with respect to both rotation and direction translation:

\begin{equation} 
\hat{\bm{R}} , \hat{\bm{u}} = \argmin_{\bm{u} , \bm{R}} \bm{u} \bm{M\left(\bm{R}, \right)} \bm{u}^\intercal
\end{equation}

The computation of the Jacobians of this equation is simplified following the derivations available in Briales et al. \cite{briales2018certifiably} to obtain the following functional:

\begin{equation} 
\hat{\bm{R}} , \hat{\bm{u}} = \argmin_{\bm{R} \in \text{SO(3)} \bm{u} \in S^2} \bm{x} \bm{C} \bm{x}^\intercal
\label{eq:final1}
\end{equation}

Where the data matrix $\bm{C} \in \text{Sym}_{27}$ gathers all data (derived from the original bearing vectors). $\bm{x} = vec\left(\bm{ru}^\intercal\right)$  where $\text{vec}()$ is a vectorization function and $\bm{r}$ is the vectorized form of the rotation matrix $\bm{R}$.

At this stage, we diverge from the QCQP procedure in Briales et al. \cite{briales2018certifiably} to solve the optimization problem. While Briales et al. \cite{briales2018certifiably} investigated the properties of the optimization (certifying the solution is globally optimal), we are interested in building a real-time VO system where the relative pose estimator is its main building block. For this reason, we choose a Levenberg–Marquardt optimizer, which is a classical optimizer that is better suited for frame-to-frame optimizations since it is efficient and allows us to leverage predictions from the previous frame as initial guesses. We show in the experimental section that our proposed estimator outperforms QCQP (Briales et al. \cite{briales2018certifiably}) when a decent initial guess is available (even if our estimator does not come with global optimality guarantees). This is because the global minimum is not always associated with an accurate 5 DoF solution and therefore an initial guess makes it easier to find an accurate solution.
The challenge with the equation \ref{eq:final1} is that the residual is scalar and therefore the number of unknowns (5) does not match the size of the residual (1). To solve this problem, we applied the trick proposed in Kneip and Lynen \cite{kneip2013direct} where instead of minimizing the functional, we minimize the Jacobians of the functional. That way, the dimension of the residual equals the number of unknowns. For clarity, we split the 5 parameters to optimize in two different variables. The $\text{so(3)}$ rotation parameters are stored in variable $\bm{\theta}$ while the 2 parameters to optimize from the 3D unit-norm manifold are stored in variable $\bm{\beta}$. Using this convention, the functional reads as follows:

\begin{equation} 
\hat{\bm{R}} , \hat{\bm{u}} = \argmin_{\bm{R} \in \text{SO(3)}, \bm{u} \in S^2} \sum_{i=1}^{3} \dfrac{\partial\bm{x} \bm{C} \bm{x}^\intercal}{\partial{\theta_i}} + \sum_{i=1}^{2} \dfrac{\partial\bm{x} \bm{C} \bm{x}^\intercal}{\partial{\beta_i}}
\label{eq:final2}
\end{equation}

The second contribution of this paper is the inclusion of the functional in the residual to optimize to further improve the results. The original residual has convergence issues when the initial guesses of the parameters to estimate are not good because it only minimizes the Jacobians of the functional. Adding the functional to the residual improves the convergence properties of the problem, as shown in the experimental section. The final residual $f(\bm{R}, \bm{u})$ reads as follows:

\begin{equation} 
\hat{\bm{R}} , \hat{\bm{u}} = \argmin_{\bm{R} \in \text{SO(3)}, \bm{u} \in S^2} f(\bm{R}, \bm{u})
\label{eq:final3}
\end{equation}
\begin{equation} 
f(\bm{R}, \bm{u}) =  \sum_{i=1}^{3} \dfrac{\partial\bm{x} \bm{C}  \bm{x}^\intercal}{\partial\theta_i}  + \sum_{i=1}^{2} \dfrac{\partial\bm{x} \bm{C} \bm{x}^\intercal}{\partial{\beta_i}} + W  \bm{x} \bm{C} \bm{x}^\intercal
\label{eq:final4}
\end{equation}

Where the constant $W$ is used to tune the properties of the estimation by differently weighting the Jacobians and 1-d residual.  How this constant influences the results is evaluated in Section \ref{new_functional}. The functional is minimized using a standard Levenberg–Marquardt optimizer where the initial guess for the rotation can be obtained by gyro propagation if available or from the previous frame otherwise. The initial guess for the translation is obtained using equation \ref{eq:t_ab} given the set of feature correspondences and the initial guess for the rotation.

The Jacobians of the functional ($\dfrac{\partial\bm{x} \bm{C} \bm{x}^\intercal}{\partial\theta_i}$ and $\dfrac{\partial\bm{x} \bm{C} \bm{x}^\intercal}{\partial\beta_i}$) are derived in the appendix of the paper.

The main advantage of relative pose estimator methods is that they always estimate an accurate rotation (even in low parallax motions) and do not need the depth of the feature matches to do so. One issue is that the translation direction might not be accurate in low-parallax motion. Fortunately, we can leverage the fact that translations are not perceivable by the user in low-parallax motions. Another issue is that we cannot use robust cost functions to down-weight outlier matches because the re-projection errors are not directly part of the residual. Because of this, we need to heavily rely on the RANSAC (Fischler et al. \cite{fischler1981random}) approach that is explained in the next Section \ref{details}.

The limitation of our relative pose estimator is its low accuracy when the initial guess for the rotation is not good. This is expected as we do a non-linear optimization where the initial guess is refined. This is quantitatively demonstrated in experiment \ref{state_of_the_art}.

\subsection{1-DoF translation magnitude estimator} 
\label{scale}

The relative pose estimator estimates unit norm translations where the magnitude ($s$) of the translation is unknown. This is addressed by this component, which estimates the magnitude of the translation ($s$) and updates the final 3-DoF translation as $\bm{t} \in R^3 :=  s  * \bm{u}  \in S^2$. 
   
The translation magnitude is optimized using a Levenberg–Marquardt solver by minimizing the squared re-projection errors (${r}_{geo}$) of the features with estimated depth, which are re-projected from the keyframe to the current frame.

\begin{equation}
\hat{s} = \argmin_{s} {r}_{geo},
\label{eq:scale}
\end{equation}

\begin{equation}
{r}_{geo} = \sum_{i=1}^{n}  g\left(\dfrac{ \left( \pi\left(\bm{R} \bm{f_i} d_i  + \bm{u} s\right) - \pi \left(\bm{f_i}^\prime \right) \right)^2} {\sigma_{i}^2}   \right)
\end{equation}

Where $g()$ is a robust cost function and $\pi()$ is the camera projection function that transforms 3D points in the camera frame into 2D camera coordinates. $d_i$ is the estimated depth of the feature in the source frame and $\sigma_{i}$ is the uncertainty of the pixel, taken from the pyramid level where the feature was observed.

\begin{figure*}[h!]
\centering
     \includegraphics[width=1.0\textwidth]{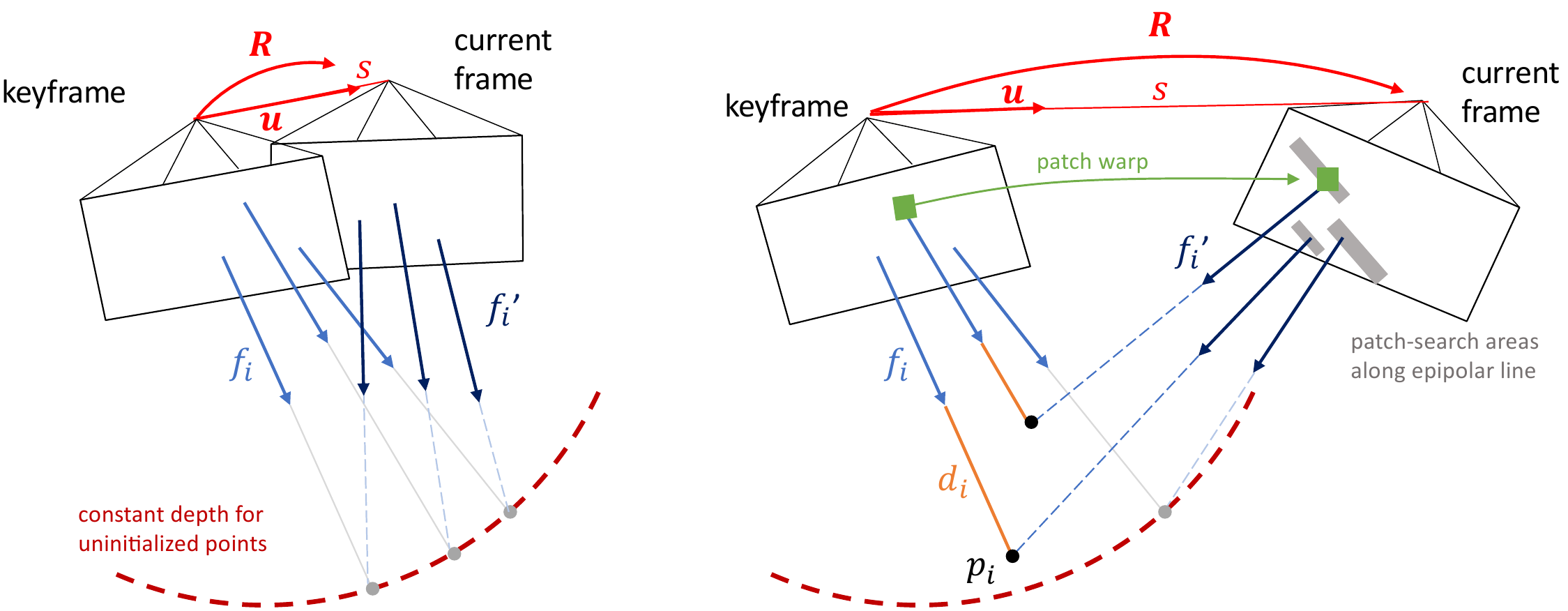}
      \caption{Left image: For the very first frames, we cannot estimate the depth of the features. All features are used for both estimating the relative pose ($\bm{R}$, $\bm{u}$) and the translation magnitude ($s$). For the translation magnitude estimation, we assume constant depth for all features. Right image: Once the baseline between cameras increases, we can estimate the depth of some of the features. Again, all features are used to estimate the relative pose  ($\bm{R}$, $\bm{u}$). However, only those features with estimated depth are used for estimating the translation magnitude $s$. Our method moves from left to right and therefore stops using the constant depth assumption once 10 features with estimated depth are available.}
       \label{fig:relpose}
\end{figure*}

\section{Implementation details}
\label{details}

\subsection{Constant depth assumption}
\label{constant_depth_assumption}
Note that for the very first frames, we do not know the depth $d_i$ of the features and we assume constant depth of 0.75 meters. As soon as we pass a minimal parallax angle (which was experimentally set to 1.0 degrees), we start estimating depth by triangulating the feature correspondences. The reader might wonder what happens during the first frames in low-parallax cases. The pose estimator estimates a relative 5-DoF pose between keyframe and frame where the rotation should always be correct and the unit-norm translation might not be correct in cases of low parallax motion. However, in such cases, the translation magnitude estimator will estimate a magnitude that will be close to zero (features are far away) and therefore the relative translation between keyframe and frame will be close to zero and will not affect negatively to the global estimation (and the user will not perceive it either). The accuracy of the translation direction and the translation magnitude naturally increase as the baseline between keyframe and frame also increases. The higher the baseline, the more perceivable the translation becomes and the more features we can triangulate. We keep using the constant depth assumption until we have a minimum set of triangulated features. The reader is referred to Figure \ref{fig:relpose} for a graphical explanation.

\subsection {Frame-to-frame tracker}
We match features from frame-to-frame by projecting 3D points (see previous subsection), corresponding to the features in the keyframe, into the current frame and correlating a warped 8x8 pixel-size patch along the epipolar line in the current view, similar to the depth estimation stage in Forster et al. \cite{forster2016svo}.  To account for pose prediction inaccuracy, the search width around the epipolar line is variable. The output of the matching stage are keyframe-to-frame feature matches ($\bm{f_i}^\prime$) which are used for relative pose estimation (section \ref{relative}).
The frame-to-frame tracker also contains a module to refine the depth ($d_i$) of the tracked features in the keyframe once the pose $\bm{T}$ is estimated (see Figure \ref{fig:pipeline}). The depth of the features is used to estimate  the magnitude of the relative translations (see Section \ref{scale}) and to update the 3D points corresponding to the features in the keyframe.

 If a feature is lost or is not observed by the tracker, we still keep it alive for 150 frames. We do this to keep features alive for as long as possible to "simulate" that a pseudo-map is tracked and our odometry can therefore re-use old information and increase the accuracy and robustness of the tracker --specially with pure rotational motions, where the pseudo-map cannot be augmented.

\subsection {Keyframe heuristics}
The 6-DoF pose estimator (5-DoF from Section \ref{relative} + 1-DoF from Section \ref{scale}) estimates the pose of the current frame with respect to a previous keyframe. Estimating the pose with respect to a keyframe, instead of the previous frame, reduces drift. 

New keyframes are added based on the following heuristics: not enough inliers in the relative pose estimator --we classify a point as an inlier if its Sampson distance (Fathy et al. \cite{fathy2011fundamental}) is small enough--, not enough overlapping features with estimated depth between keyframe and frame and high re-projection error in the translation magnitude estimator. If a keyframe has to be inserted, it is inserted in the previous frame and we estimate the 6-DoF pose between the last 2 frames. Note that at the time we do pose estimation, we don't know the depth of the features in the current frame. However, since we do a keyframe to frame optimization we do know the depth of the features in the keyframe as those were triangulated when the keyframe was processed.

\subsection{Outlier removal}
As with all optimization problems, the formulation of the relative pose estimator (section \ref{relative}) is sensitive to outlier feature tracks. However, we are not dealing with a per-feature geometric error but an algebraic one, meaning we cannot use robust cost functions to down-weight outliers. Hence, we wrap the minimization in a RANSAC loop (Fischler et al. \cite{fischler1981random}) where first we initialize the relative rotation from the best estimated rotation so far. If the rotation have not been estimated yet, we use the rotation prior from gyro prediction if available. If a rotation prior is not available, we apply a small perturbation to the estimated rotation from previous frame. The translation direction is initialized from equation \ref{eq:t_ab} using the  subsampled --from the inliers set-- feature correspondences. We repeat this operation during 5 iterations, updating the inliers of our problem.
After the 5 iteration, we additionally refine the pose (minimizing eq. \ref{eq:final3}) during 7 more iterations. After every iteration, outliers are always removed with respect to the entire set of correspondences. To determine if a point is an outlier we check if its squared Sampson distance is higher than a predefined threshold. We stop early if we reach convergence, which is obtained if the error is not reduced and the number of inliers does not increase. Otherwise, we update the best relative pose and the inliers.

The translation magnitude estimator uses only the inliers coming from the relative pose estimator. We use a robust cost function to down-weight outliers. After the estimation, we detect additional outliers by checking the re-projection error of the features. A feature is considered an outlier if the re-projection error is higher than 1.5 pixels. Outliers from both the relative pose estimator and the translation magnitude estimator are removed from the tracker in a post-processing step.

\subsection{Initial guess for translation magnitude estimator}
 
When we introduce a new keyframe in the previous frame ($n-1$), the translation magnitude is computed with respect to that frame. In this case, the estimated translation magnitude will be very close to zero, and therefore we initialize the initial guess to zero ($s_n = 0$). Otherwise, when we estimate the translation magnitude with respect to a keyframe that is older than the previous frame, the current magnitude ($s_{n}$) is initialized from the last estimated magnitude  ($s_{n} := \hat{s}_{n-1}$). However, we need to take into account that the magnitude sign may flip from frame to frame. This is because the initial guess for the translation is computed feeding the initial guess of the rotation into the closed form solution from equation \ref{eq:t_ab}, which can flip the sign of the translation due to its symmetry in the functional of this equation ($\bm{u} \bm{M\left(\hat{\bm{R}}\right)} \bm{u}^\intercal = \left(-\bm{u}\right) \bm{M\left(\hat{\bm{R}}\right)} \left(-\bm{u}^\intercal\right)$). To solve this, we check if the sign of the 2-DoF direction has flipped from last frame ($\bm{u_{n}} \simeq - \hat{\bm{u}}_{n-1}$) and if that is the case we also flip the sign of the magnitude prediction ($s_{n} := - \hat{s}_{n-1}$).


\section{Experiments}
\label{experiments}

\subsection{Pose estimator validation}
\label{pose_estimator_validation}

In this section, we compare our proposed pose estimator (5-DoF (eq. \ref{eq:final3}) + 1-DoF (eq. \ref{eq:scale})) against the gold standard solution for this problem, which is doing Bundle Adjustment (Triggs et al. \cite{triggs1999bundle}) without optimizing the position of the 3D points and therefore only optimize the 6-DoF pose of the camera:

\begin{equation}
\hat{\bm{R}},  \hat{\bm{t}} = \argmin_{ \bm{R \in \text{SO(3)}} ,   \bm{t \in  \mathbb{R}^3 }  } r,
\end{equation}

\begin{equation}
r = \sum_{i=1}^{n}  g \left( \dfrac{\left( \pi\left(\bm{R} \bm{f_i} d_i  + \bm{t} \right) - \pi \left(\bm{f_i}^\prime \right)  \right)^2}{\sigma_{i}^2} \right)
\end{equation}

We simulate a map of 200 landmarks and a trajectory of $\sim$1 second (37 frames) with a sigma equal to 0.75 for the pixel uncertainty. Around 170 landmarks are observed per frame and from a distance to first camera between 1 and 6 meters. The camera moves around  25 degrees and 1.0 meters on average. We use a spherical camera model (as in Kneip and Lynen \cite{kneip2013direct}) with an image size of 640 by 480 and a focal length of 200 pixels. As stated by Briales et al. \cite{briales2018certifiably}, increasing the Field Of View (FOV) makes the relative pose problem easier, as this results in a better constraining of the optimization objective. The reader is referred to Briales et al. \cite{briales2018certifiably} for an evaluation of the FOV in the relative pose problem.

We run 50 experiments for each algorithm. In this experiment, the estimation is always computed with respect to the first frame and keyframes do not need to be inserted since most of the landmarks are observed by the 37 frames. This ensures a fair comparison since we focus on evaluating the pure estimation and not the keyframing heuristics. We are interested in two comparisons:
\begin{itemize}
\item Comparing both approaches when the depth is unknown. We use the constant depth assumption and therefore fix all depths to the same value.  This is to validate our proposal, showing it is particularly accurate during the first frames when depth is not available and 6-DoF approaches suffer.
\item Comparing both approaches when the depth of the features is known, which is the normal use case for 6-DoF approaches. This is to confirm that our estimator works well in the normal case, meaning it is not only good for initialization but also for tracking.
\end{itemize}

\begin{figure}[ht!]
\begin{subfigure}
\centering
\includegraphics[width=0.460\textwidth]{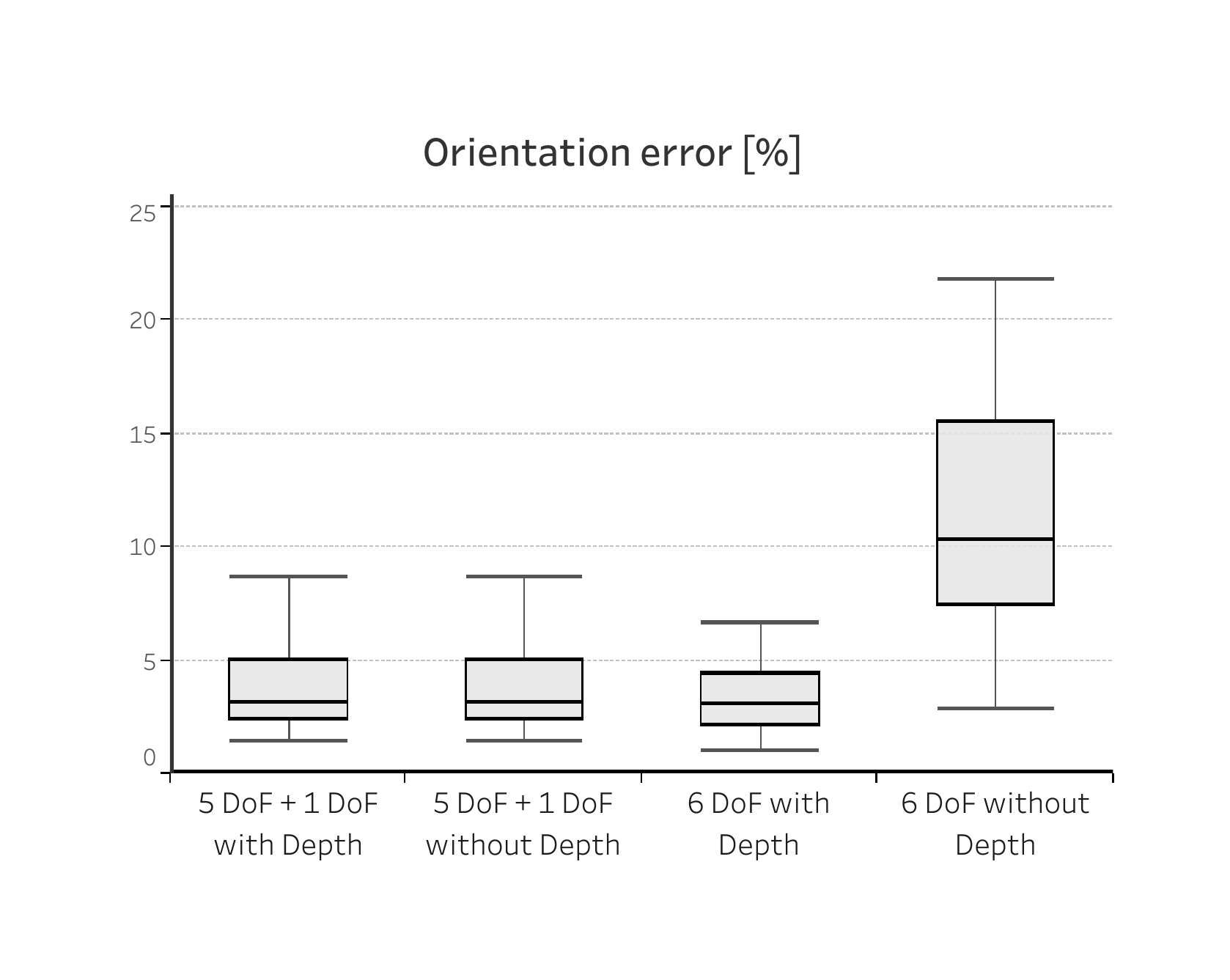}
\label{fig:estimatorValidation2}
\end{subfigure}
\begin{subfigure}
\centering
\includegraphics[width=0.460\textwidth]{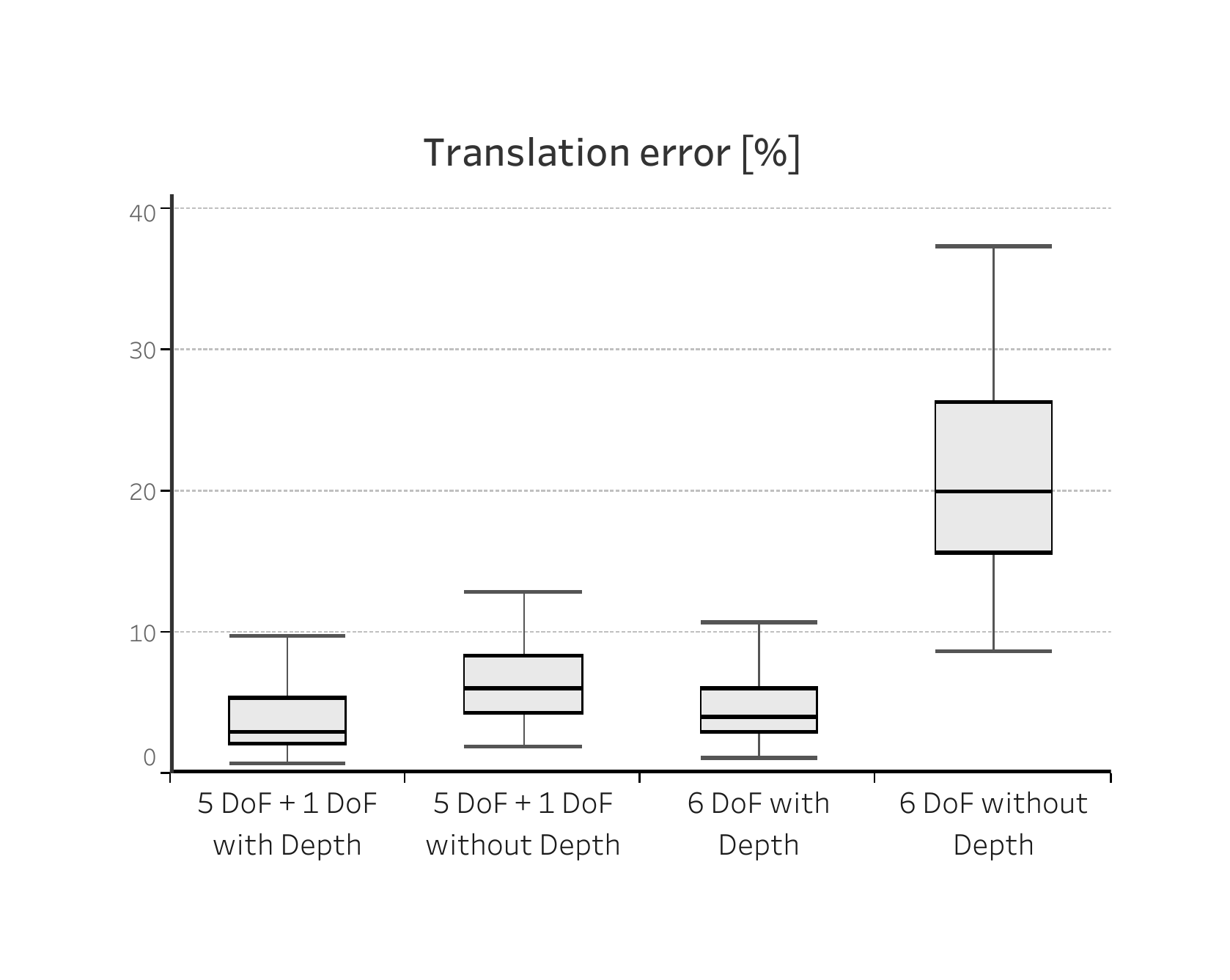}
\label{fig:estimatorValidation1}
\end{subfigure}
\caption{Comparison between classical 6-DoF estimators and our proposal (5-DoF + 1-DoF estimator). We distinguish between two cases: 1) general case, where depth is available and 2) depth is not available -- i,e. during initialization phase. Errors are reported as the ratio  [$\%$] between the maximum error (orientation or translation) and the maximum displacement (rotation or translation) between any two points in the trajectory. Our approach outperforms the classical 6-DoF estimator when depth is not available (more details in the text). The whisker plots representation is as follows: P5, P25, median, P75 and P95.}  
\label{fig:estimatorValidation}
\end{figure}

The problem of the 6-DoF estimator is that the error in the estimated depth of the features is going to propagate to all the estimated parameters (rotation and translation). However, in our (5-DoF + 1-DoF) estimator, the error in the estimated depth only propagates to the translation magnitude and it does not have any influence in the estimated rotation or in the translation direction. Our results, that can be observed in Figure \ref{fig:estimatorValidation2}, confirm these hypotheses. Note that the rotational error in our approach is independent of the accuracy of the depth, which is really important for our use case as users might rotate their phones without applying any translational motion. On the other hand, the 6-DoF estimator estimates a significantly less accurate rotation if the estimated depth is unknown, confirming our hypothesis. 

In Figure \ref{fig:estimatorValidation1} we can withdraw a similar conclusion for the translation error. In this case, if the depth is unknown our estimator is also more accurate than the 6-DoF estimator. When the depth is known, both estimators behave similarly. Also, as expected, the final translation error is increased in both estimators when depth is unknown. However, not having depth has a way smaller influence in our estimator (median error increases from 3 $\text{\%}$ to 6 $\text{\%}$) than in the 6-DoF estimator (median error increases from 3 $\text{\%}$ to 19 $\text{\%}$). This is because depth error only propagates to the translation magnitude in our approach while it propagates to all estimated parameters in the 6-DoF estimator.

\begin{figure}[bt]
\centering
\includegraphics[width=0.460\textwidth]{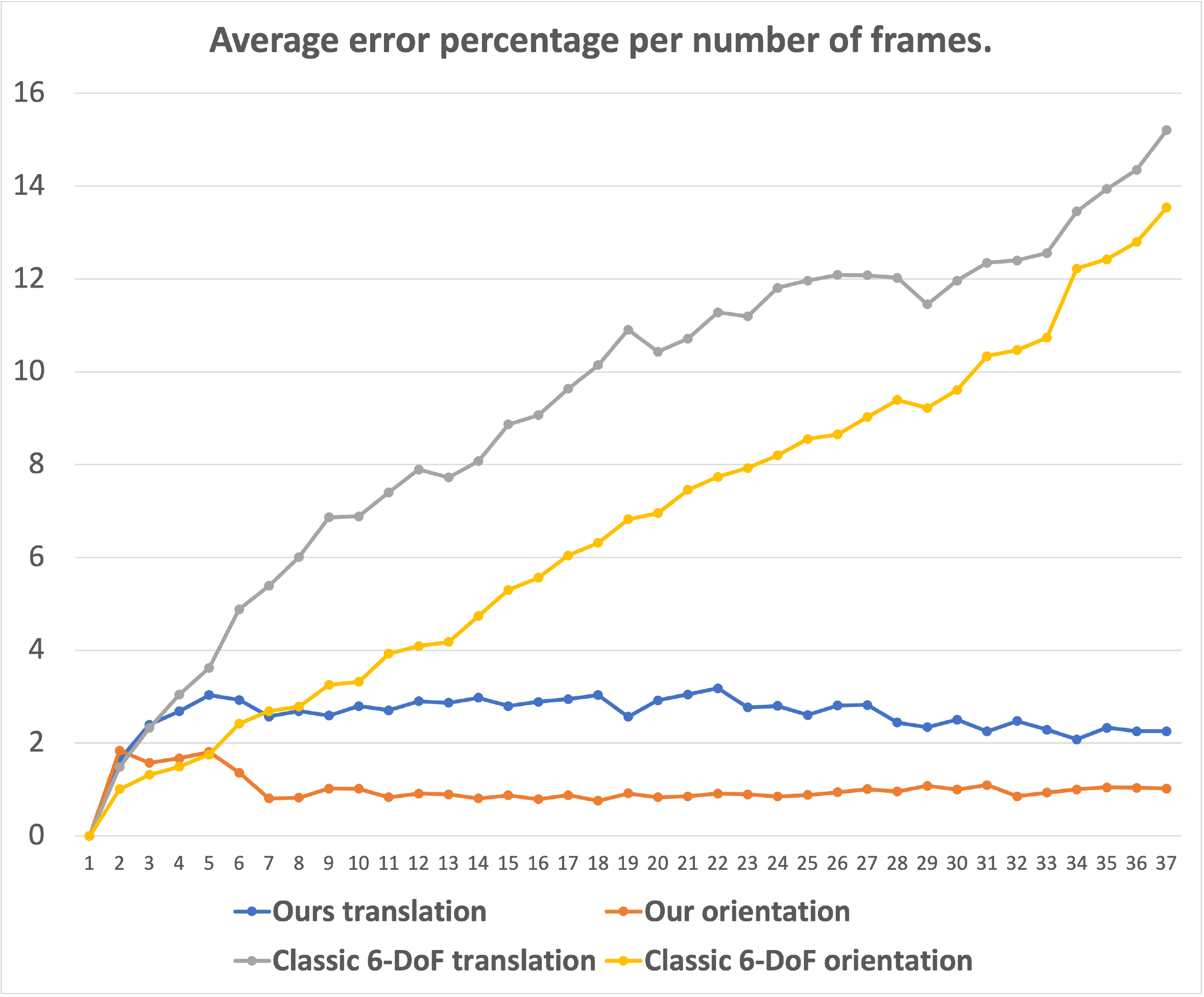}
\caption{Orientation and translation error per number of frames. We compare our approach against the classic 6-DoF estimator when depth is not available. Out approach has a bounded average error while the error of the classic 6-DoF estimator linearly increases with the frame number. Errors are reported as the ratio  [$\%$] between the average absolute error in a frame and the maximum displacement (orientation or rotation) between any two points in the trajectory. }
\label{fig:error_per_frame}
\end{figure}

Using the same data from previous experiment, we also report the accumulated keyframe-to-frame average errors per number of frames in Figure \ref{fig:error_per_frame}. This is a very important result as it shows our approach has a low and bounded error for the entire trajectory (between 1 $\%$ and 3 $\%$) while the classic 6-DoF estimator accuracy linearly deteriorates as the baseline (number of frames) increases. The final error for the classic estimator is one order of magnitude worse after 37 frames (between 13 $\%$ and 15 $\%$). 

These experiments confirm the validity of our approach, demonstrating that it can be a good replacement for the classical 6-DoF estimator thanks to the fact it has better properties, specially during initialization where we face low-parallax motions and features cannot be triangulated accurately.

The interested reader can also confirm the validity of our approach by looking at our video from the supplementary material (\url{https://youtu.be/ZGmzXK-dj1Y}).

\subsection{Comparison against the state-of-the-art}
We have compared our approach against state-of-the-art approaches for the relative pose problem. The TUM benchmark has been used for the evaluation.
\label{state_of_the_art}

\subsubsection{Dataset for the relative pose problem}
\label{dataset}

We release a dataset that we have generated from TUM benchmark (Sturm et al. \cite{sturm2012benchmark}) to evaluate relative pose estimators and therefore facilitate the comparison against future solutions for the relative pose problem. To this purpose, we use six recordings from TUM benchmark -- two from each different \emph{Freiburg} Kinect sensor. We run ORB-Slam (Mur-Artal et al. \cite{mur2017orb}) in every recording and we store the keyframe-to-frame bearing correspondences --after removing potential outliers with our RANSAC approach-- into text files. The corresponding ground-truth pose from keyframe-to-frame is also provided for quantitative evaluation. We also generate the same feature correspondences but without noise to facilitate the debugging when first using this dataset. We produce the noiseless correspondences by projecting the 3D map points associated to those correspondences into the keyframe and the frame. We subsample the dataset to have an affordable number of keyframe-to-frame pairs per recording, resulting in around 300 pairs per recording on average. Both the dataset and the results obtained in this paper for each algorithm (see next Section) can be downloaded from \url{https://github.com/facebookresearch/relative_pose_dataset}.


\begin{figure*}[bt!]
\centering
\includegraphics[width=1.0\linewidth]{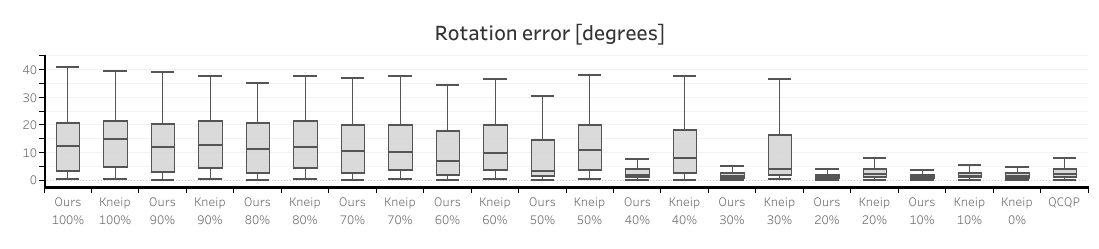}
\caption{Comparison between QCQP approach from Briales et al. \cite{briales2018certifiably}, Kneip and Lynen \cite{kneip2013direct} and our approach. We have used the TUM dataset "fr3 structure texture near" for this comparison. The percentage number specifies the error of the initial guess (see equation \ref{eq:guess} for the definition of the initial guess error). Our approach consistently outperforms Kneip and Lynen \cite{kneip2013direct} and it also outperforms QCQP if the initial guess error is $30\%$ or smaller.  The whisker plots representation is as follows: P5, P25, median, P75 and P95. We release this dataset and our results to facilitate the comparison against other approaches in \url{https://github.com/facebookresearch/relative_pose_dataset}.}
\label{fig:near}
\end{figure*}

\begin{figure*}
\centering
\includegraphics[width=1.0\linewidth]{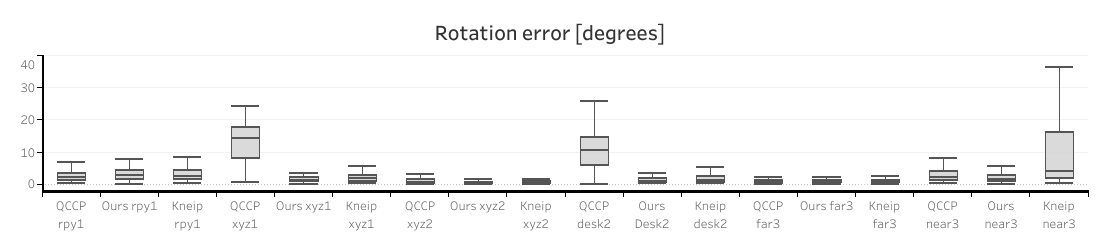}
\caption{Comparison between QCQP approach from Briales et al. \cite{briales2018certifiably}, Kneip and Lynen \cite{kneip2013direct} and our approach for 6 datasets from TUM benchmark (Sturm et al. \cite{sturm2012benchmark}). Our approach outperforms both QCQP approach from Briales et al. \cite{briales2018certifiably} and Kneip and Lynen \cite{kneip2013direct} in most of the recordings. We have used an initial guess error of  $30\%$ in this experiment. See equation \ref{eq:guess} for the definition of the initial guess error. The whisker plots representation is as follows: P5, P25, median, P75 and P95. Legend of the datasets is as follows: rpy1 (fr1 rpy),  xyz1 (fr1 xyz), xyz2 (fr2 xyz), desk2 (fr2 desk2), far3 (fr3 structure texture near) and near3 (fr3 structure texture far). We release this dataset and our results to facilitate the comparison against other approaches in \url{https://github.com/facebookresearch/relative_pose_dataset}.}
\label{fig:all}
\end{figure*}

\subsubsection{Evaluation with TUM dataset \cite{sturm2012benchmark}}
\label{tum_dataset_evaluation}

In this section, we compare our proposal for the relative pose estimator problem against the OpenGV implementation (\url{https://laurentkneip.github.io/opengv/}) provided by Kneip and Lynen \cite{kneip2013direct} and QCQP solution from Briales et al. \cite{briales2018certifiably}. QCQP code is not available online but it was provided by the authors. We use the dataset that we explained in the previous subsection \ref{dataset} for this evaluation. 

Current implementation of the QCQP solution from Briales et al. \cite{briales2018certifiably} does not admit initial guesses while our proposal and Kneip and Lynen \cite{kneip2013direct} proposal do admit initial guesses. As a first experiment, we evaluate the importance of the initial guesses. We compare the three approaches in a recording from TUM benchmark (Sturm et al. \cite{sturm2012benchmark}) and report the results we obtained with different accuracy levels of the initial guesses.

The initial guesses are generated as follows:

\begin{equation}
\label{eq:guess}
R_\text{guess} = \exp_{so(3)}\left({\log_{SO(3)}(R_\text{gt}) * (1.0 - \Gamma)}\right)
\end{equation}

Where $R_\text{gt}$ is the ground-truth relative pose between both cameras and $\Gamma$ goes from 0.0 (perfect initial guess) to 1.0 (inaccurate initial guess). In our figures, an initial guess error of $X\%$ means $\Gamma = 0.01 X$ in this equation.

As it can be observed in Figure \ref{fig:near}, our approach and Kneip and Lynen \cite{kneip2013direct} approach are quite inaccurate when the initial guesses are not good. This is expected as these approaches are non-linear optimizations that normally find a local minimum close to the initial guess. QCQP solution from Briales et al. \cite{briales2018certifiably} principal advantage is that is a solution that comes with global optimality guarantees making this solution the preferred one, specially if initial guesses are inaccurate. However, the global minimum is not guaranteed to be the one that achieves the most accurate pose, specially in ill-posed problems. For this reason, Kneip and Lynen \cite{kneip2013direct} and our approach obtain comparable or even better results than QCQP approach from Briales et al. \cite{briales2018certifiably} if the initial guesses are good enough. According to these experiments, our approach can handle initial guess errors of up to $30\%$, which is good enough as the initial guesses are computed from the previous frame, meaning the potential $\Gamma$ we need to handle is equal to $1 - (N - 1) / N$ and therefore the initial guess error will to zero as $N$ increases. Where $N$ is the number of frames between keyframe and frame.
Note that our approach has better convergence properties than Kneip and Lynen \cite{kneip2013direct}, it is able to converge to an accurate solution using worse initial guesses.  This is mostly due to our second contribution of the paper, which is combining the functional and its Jacobians in the residual which helps finding the global minimum easier. We analyze this weight in more detail in the Section \ref{new_functional}.

In a second experiment (see Figure \ref{fig:all}), we compare the three approaches in six recordings from the TUM benchmark. In this experiment, we have used the initial guesses with $30\%$ of error. Note that our approach outperforms both QCQP approach from Briales et al. \cite{briales2018certifiably} and Kneip and Lynen \cite{kneip2013direct} in most of the recordings.



\subsection{Validation of the proposed residual}
\label{new_functional}
The previous experiment from Section \ref{state_of_the_art} validates our proposal for the residual of the iterative relative pose problem. However, we include a second experiment to further analyze the influence of our contribution in the formulation. The influence of our proposal can be tuned with the weight $W$. For this experiment, we use synthetic data and same camera model as in the experiments from Section \ref{pose_estimator_validation}.  We create a random set of 2D-2D correspondences between two different camera views (with uncertainty equal to 0.75 pixels).  We perturb the poses, and the initial guesses for the rotation and translation direction contain an error of around 10 degrees and 50 degrees in average, respectively. We run the relative pose estimator and compute the error as a function of the weight $W$ (which ranges from 0 to 1000). We run 50 experiments per weight.

\begin{figure}[bt]
\centering
\includegraphics[width=0.460\textwidth]{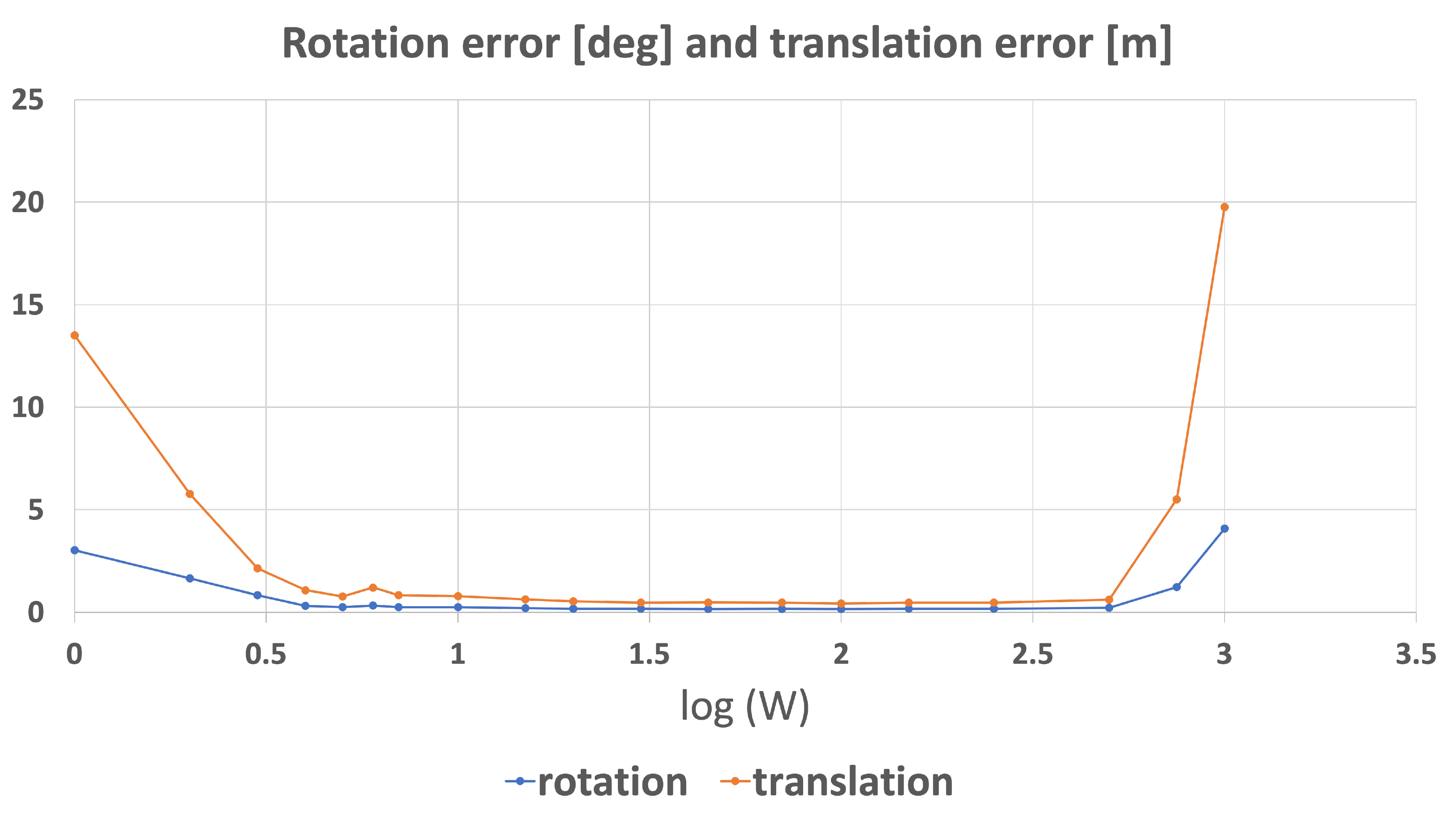}
\caption{Positional and rotational errors as a function of the weight $W$ proposed in equation \ref{eq:final3}. The Figure shows that our proposal does improve the results significantly in optimizations where the initial guess is far from the actual solution. The improvements are up to a factor of 10. Also, the estimator is quite robust to this weight, values between 15 and 250 obtain a similar level of accuracy.}
\label{fig:weight}
\end{figure}

Observe in Figure \ref{fig:weight} that the best results are obtained with weights bigger than zero. However, note that we get big errors if we use huge values for the weight. In this case, the problem becomes ill-posed as the functional (1-D) will have a way higher influence than the Jacobians (5-D) and the residual will become effectively 1-D while the number of parameters to estimate is still 5. Having a weight equal to zero results in the original residual (with only the Jacobians), resulting in high errors in average when compared against our approach. 

The new residual is particularly helpful when the initial guess of the relative pose is far from the optimal one. In this case, the convergence properties are improved as the functional is included in the residual. As expected, if one repeats this test with good initial guesses for the relative pose, the results obtained for both residuals will be accurate and very close to each other. 

\begin{figure*}[bt!]
\centering
\includegraphics[width=1.0\linewidth]{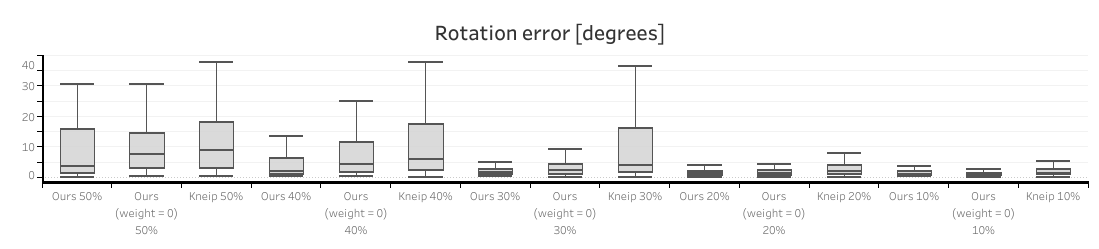}
\caption{Evaluation of the addition of the functional in the residual. Experiment using TUM dataset "fr3 structure texture near". The percentage number indicates the initial guess error (see equation \ref{eq:guess}). Note our results are worse if the functional is not included in the residual (weight = 0). Our approach can converge for higher errors of the initial guess when compared against Kneip and Lynen \cite{kneip2013direct} approach. The whisker plots representation is as follows: P5, P25, median, P75 and P95.}
\label{fig:weight_tum}
\end{figure*}

We do an additional experiment with real data to further validate the inclusion of the functional in the residual. The reader is referred to Figure \ref{fig:weight_tum} for more details.

\subsection{Complexity of our approach}
Our pipeline runs in real-time in both low-end and high-end devices. We have measured the compute of our full visual odometry (tracker and pose estimator) in a Samsung-S10 and in a Huawei-P20. We have obtained average compute times of $\sim$10 and $\sim$25  ms respectively. The frame to frame tracker and the pose estimator represent the $\sim60\%$ and $\sim40\%$ of the total compute respectively. While Kneip and Lynen \cite{kneip2013direct} estimator also works in real time and has a similar complexity as our approach, QCQP approach from Briales et al. \cite{briales2018certifiably} is implemented in \emph{Matlab} and it takes several seconds of compute per frame.

\subsection{Failure cases.}
As demonstrated in the experimental section, the main failure case of our approach is the low accuracy achieved when the initial guess is far from the true value. Even though we tried to mitigate this issue by robustifying our residual, we still have convergence issues and working on a more robust solution remains for future work. 

Similarly to most relative pose estimators in the literature, our relative pose estimator is very sensitive to outliers. Even though we do not handle outliers in the estimator itself, we wrap the relative pose estimator inside of a RANSAC scheme to mitigate this issue.

\section{Conclusions and future work}
\label{conclusion}

In this paper, we have proposed a novel monocular visual odometry that initializes without motion parallax and estimates a 6-DoF pose from first frame on. This is achieved by combining a relative pose estimator and a translation magnitude estimator. We have shown this estimator outperforms the classical 6-DoF estimator in initialization stages with low parallax motion. We believe the usage of relative pose estimators has a great potential in monocular SLAM due to their ability of not failing with pure rotational motions or with poor map estimation. For the relative pose problem, we have also proposed a new residual combining the Jacobians of the functional and the functional itself. The residual is minimized using a Levenberg–Marquardt optimizer. We demonstrate that minimizing both errors is more accurate than minimizing only the Jacobians.

There are some opportunities for future work:
\begin{itemize}
 
\item finding an analytic solution for the introduction of feature uncertainties and robust cost functions to down-weight outliers in the functional to minimize in the relative pose estimator.

\item merging the translation magnitude and relative pose estimators into a single probabilistic estimation module that can consume both features with depth and without depth instead of decoupling the estimation in two different modules.
\end{itemize}


\section{Appendix}
\label{appendix}

In this section, we derive the Jacobians of the functional with respect to the $\text{so(3)}$ rotation parameters $\dfrac{\partial\bm{x} \bm{C} \bm{x}^\intercal}{\partial\theta_i}$ \normalsize and the translation direction 2-DoF manifold $\dfrac{\partial\bm{x} \bm{C} \bm{x}^\intercal}{\partial\beta_i}$ that are used as part of the residual to minimize in equation \ref{eq:final3}.

We apply the chain rule to derive them. First, we compute the derivative with respect to $\bm{x}$:

\begin{equation}
\dfrac{\partial\bm{x} \bm{C} \bm{x}^\intercal}{\partial{\bm{x}}} = 2  \bm{C}  \bm{x}
\end{equation}

The Kronecker product (Van \cite{van2000ubiquitous}) is used to get the derivative with respect to the translation direction $\bm{u}$ and the vectorized form of the rotation ($\bm{r} = \text{vec}(\bm{R})$).

\begin{equation}
\dfrac{\partial{\bm{x}}}{\partial{\bm{r}}} = \text{kroneckerProduct}\left( \bm{u}, \bm{I_{9x9}}\right)
\end{equation}
\begin{equation}
\dfrac{\partial{\bm{x}}}{\partial{\bm{u}}} = \text{kroneckerProduct}\left(\bm{I_{3x3}}, \bm{r}\right)
\end{equation}

 Relative of the rotation matrix with respect to each axis $i = {x, y ,z}$:
 
\begin{equation}
\dfrac{\partial{\bm{r}}}{\partial{\theta_i}} = \text{vec}\left( \bm{R}  {[\text{unit}_i]}_x \right)
\end{equation}
$[]_x$ is the skew-symmetric matrix and $\bm{\text{unit}}_i$ is the unit vector in the direction of the $i$ axis.
 
 The translation direction is parametrized using the z-axis (third column) of a rotation matrix ($\bm{R_u} \in SO3()$ ). Therefore, its Jacobian is computed as follows:
 
\begin{equation}
\dfrac{\partial{\bm{u}}}{\partial{\bm{\beta}}} = \text{vec}\left( \left[\bm{R_u} {[\text{unit}_z]}_x \right]_{\text{topBlock(3,2)}} \right)
\end{equation}

The final Jacobians are computed applying the chain rule:

\begin{equation}
\dfrac{\partial\bm{x} \bm{C} \bm{x}^\intercal}{\partial{\theta_i}} =  \dfrac{\partial\bm{x} \bm{C} \bm{x}^\intercal}{\partial{\bm{x}}} \dfrac{\partial{\bm{x}}}{\partial{\bm{r}}}  \dfrac{\partial{\bm{r}}}{\partial{\theta_i}}
\end{equation}
\begin{equation}
\dfrac{\partial\bm{x} \bm{C} \bm{x}^\intercal}{\partial{\bm{\beta}}} =  \dfrac{\partial\bm{x} \bm{C} \bm{x}^\intercal}{\partial{\bm{x}}} \dfrac{\partial{\bm{x}}}{\partial{\bm{u}}}  \dfrac{\partial{\bm{u}}}{\partial{\bm{\beta}}}
\end{equation}


\bibliographystyle{abbrv-doi}

\bibliography{template}
\end{document}